\documentclass[conference]{IEEEtran}
\IEEEoverridecommandlockouts
\usepackage{cite}
\usepackage{amsmath,amssymb,amsfonts}
\usepackage{algorithmic}
\usepackage{graphicx}
\usepackage{textcomp}
\usepackage{subcaption}
\usepackage{subfloat}
\usepackage{threeparttable}
\usepackage{xcolor}
\usepackage{multirow}
\usepackage{booktabs}
\def\BibTeX{{\rm B\kern-.05em{\sc i\kern-.025em b}\kern-.08em
    T\kern-.1667em\lower.7ex\hbox{E}\kern-.125emX}}
\begin{document}

\title{DouZero+: Improving DouDizhu AI by Opponent Modeling and Coach-guided Learning \\
}
\author{Youpeng Zhao,
		Jian Zhao,
		Xunhan Hu,
		Wengang Zhou,
		Houqiang Li,
		\thanks{Y. Zhao, J. Zhao, X. Hu, W. Zhou and H. Li are with the CAS Key Laboratory of Technology in Geo-spatial Information Processing and Application System, Department of Electronic Engineering and Information Science, University of Science and Technology of China, Hefei 230027, China.}
		\thanks{W. Zhou and H. Li are also with Institute of Artificial Intelligence, Hefei Comprehensive National Science Center, Hefei, China}
}
\maketitle

\begin{abstract}
Recent years have witnessed the great breakthrough of deep reinforcement learning (DRL) in various perfect and imperfect information games. Among these games, DouDizhu, a popular card game in China, is very challenging due to the imperfect information, large state space, elements of collaboration and a massive number of possible moves from turn to turn. Recently, a DouDizhu AI system called DouZero has been proposed. Trained using traditional Monte Carlo method with deep neural networks and self-play procedure without the abstraction of human prior knowledge, DouZero has outperformed all the existing DouDizhu AI programs. In this work, we propose to enhance DouZero by introducing opponent modeling into DouZero. Besides, we propose a novel coach network to further boost the performance of DouZero and accelerate its training process. With the integration of the above two techniques into DouZero, our DouDizhu AI system achieves better performance and ranks top in the Botzone leaderboard among more than 400 AI agents, including DouZero.

\end{abstract}


\section{Introduction}
During the development of artificial intelligence, games often serve as an important testbed as they are good abstraction of many real-world problems, and more objective compared to environments specially designed for testing AI since games are developed for humans. In recent years, significant progress has been made in solving perfect-information games such as Go \cite{silver2016mastering, silver2017mastering, cazenave2021improving}, Shogi (Janpanese chess) \cite{silver2018general} and even fighting game \cite{kim2020mastering}. The current research efforts are turning to more challenging imperfect information games (IIG) where agents may cooperate or compete with each other under a partially observable environment. Encouraging achievements have been made from two-player games, such as simple Leduc Hold'em and limit/no-limit Texas Hold'em \cite{zinkevich2007regret, heinrich2016deep, moravvcik2017deepstack, brown2018superhuman} to multi-player games, including multi-player Texas Hold'em \cite{brown2019superhuman}, StarCraft \cite{vinyals2019grandmaster}, DOTA \cite{berner2019dota} and Japanese Mahjong \cite{li2020suphx}.

\begin{figure}[t]
	\centering
	\includegraphics[width=0.98\columnwidth]{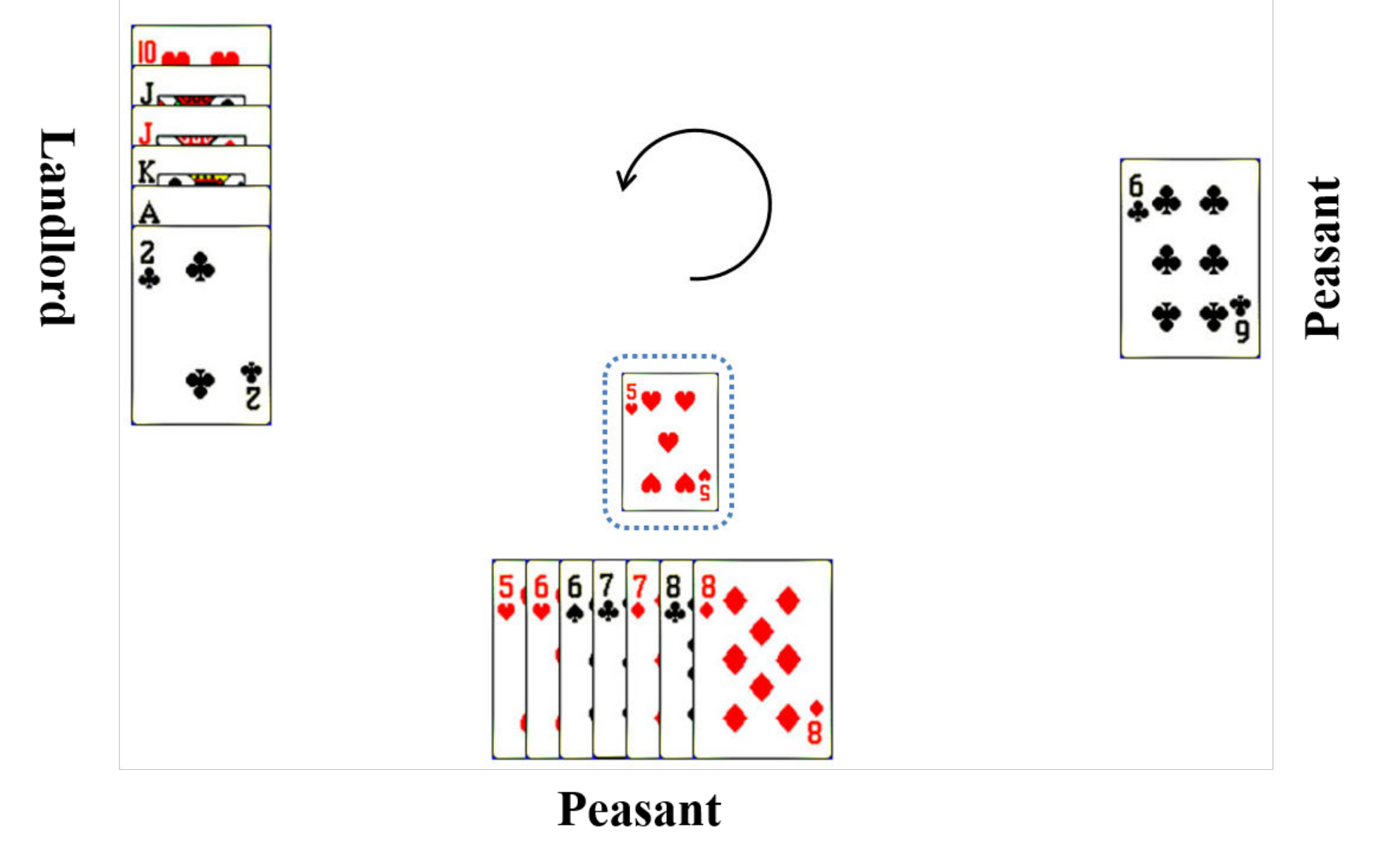}
	\caption{A case example about cooperation in DouDizhu. If the Peasants learn to cooperate with each other, current player should play small Solo to let the teammate to win the game. }
	\label{fig1}
\end{figure}

\begin{figure}[t]
	\centering
	\includegraphics[width=0.98\columnwidth]{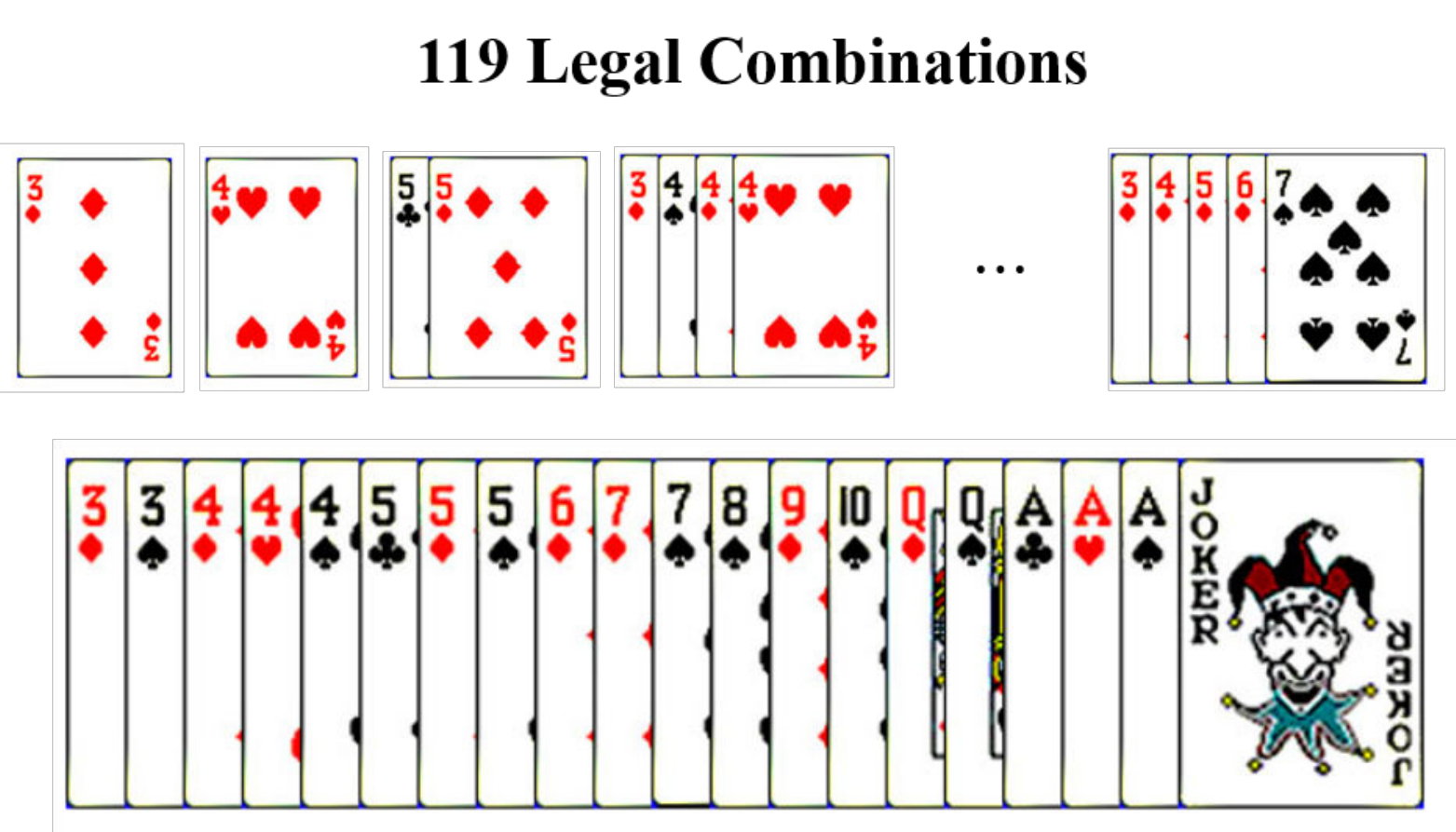}
	\caption{A hand and its corresponding legal moves.}
	\label{fig2}
\end{figure}

In this work, we are dedicated to designing an AI program for DouDizhu, \emph{a.k.a,} Fighting the Landlord, which is the most popular card game in China with hundreds of millions daily active players. DouDizhu has two interesting characteristics that pose great challenges for AI programs. First, this game involves both cooperation and competition simultaneously in a partially observable environment. To be specific, the two Peasant agents play as a team to fight against the Landlord agent. For example, Figure~\ref{fig1} shows a typical situation where the bottom Peasant can play a small Solo card to help his partner to win. This property makes the popular algorithms for Poker games, such as Counterfactual Regret Minimization (CFR) \cite{neller2013introduction} and its variants not suitable in such a complex three-player setting. Second, DouDizhu has a large and complex state and action space due to the combination of cards and the complex rules compared to other card games. There are thousands of possible combinations of cards where different subsets of these combinations are legal to different hands. Figure~\ref{fig2} exhibits an example of a hand that has 119 legal moves, including Solo, Pair, Trio, Chain of Solo and so on. Unlike Texas Hold'em, the actions in DouDizhu can not be easily abstracted, which makes search computationally expensive and the commonly used reinforcement learning (RL) algorithms less effective. The performance of Deep Q-Learning (DQN) \cite{mnih2015human} will be greatly affected due to the overestimating issue in large action space \cite{zahavy2018learn} while policy gradient methods such as A3C \cite{mnih2016asynchronous} fail to leverage the action features, limiting the capability of generalizing over unseen actions. In this way, previous work has shown that DQN and A3C have a poor performance in DouDizhu, only having less than 20$\%$ winning percentage against simple rule-based agents even with twenty days of training \cite{you2020combinatorial}.

Despite the challenges mentioned above, some achievements have been made in building DouDizhu AI. To deal with the large action space in DouDizhu, Combinatorial Q-Network (CQN) \cite{you2020combinatorial} decouples the actions into decomposition selection and final move selection. However, the decomposition selection relies on human heuristics and is time-consuming, which limits its performance. In fact, CQN does not have preponderance over the heuristic rule-based model. DeltaDou \cite{jiang2019deltadou} is the first bot that reaches top human-level performance compared to human experts. It makes use of an AlphaZero-like algorithm, which combines neural networks with Fictitious Play Monte Carlo Tree Search (FPMCTS), and an inference algorithm in a self-play procedure. However, DeltaDou pre-trains a kicker network based on heuristic rules to reduce the action space size, which may have a negative impact on its strength if the output of the kicker network is not optimal. Moreover, the inference and search are so computationally expensive that it takes two months to finish the training of DeltaDou. Recently, DouZero \cite{zha2021douzero} has attracted considerable attention due to its outstanding performance in this complex game. It utilizes self-play deep reinforcement learning without the abstraction of state/action space and human knowledge.  It combines classical Monte-Carlo methods \cite{sutton2018reinforcement} with deep neutral networks to handle the large state and action space, which opens another door for such complex and large-scale games.

In this work, we improve DouZero by introducing opponent modeling and coach-guided learning. Opponent modeling aims to determine a likely probability
distribution for the opponents’ hidden cards, which is motivated by the fact that human players will try to predict the opponents' cards to help them determine the policy. Due to the complexity of DouDizhu, a lot of actions may be appropriate when making the decision. In this case, analyzing the opponents' cards will be of great importance because grasping this information helps the bot choose the optimal move. On the other hand, we propose coach-guided learning to fasten the training of the AI. Due to the large information space in this game, the training of the AI program for DouDizhu costs a lot of time. Considering the fact that the outcome of DouDizhu depends heavily on the initial cards of three players, quite a few games are of little value for learning. To this end, we design a novel coach network to evenly pick matched openings so that the models can learn from more valuable data without wasting time to play valueless games, thus accelerating the training process. Through integrating these techniques into DouZero, our DouDizhu AI program achieves a better performance than the original DouZero and ranks the first on the Botzone \cite{zhou2017botzone, zhou2018botzone, zhang2012botzone} leaderboard among more than four hundred agents, including DouZero. 

\section{Related Work}
In this section, we briefly introduce the application of reinforcement learning in imperfect-information games as well as the works about opponent modeling.

\subsection{Reinforcement Learning for Imperfect-Information Games}
Recent years have witnessed the successful application of reinforcement learning in some complex imperfect-information games. For instance, there are quite a few works about reinforcement learning for poker games \cite{heinrich2016deep, sweeney2012applying, lanctot2017unified}. Different from  Counterfactual Regret Minimization (CFR) \cite{neller2013introduction} that relies on game-tree traversals, RL is based on sampling so that it can easily generalize to large-scale games. In this way, OpenAI, DeepMind and Tencent have utilized this technique to build their game AI in DOTA \cite{berner2019dota}, StarCraft \cite{vinyals2019grandmaster} and Honor of Kings \cite{ye2020mastering}, respectively and acquired amazing achievements, proving the effectiveness of reinforcement learning in imperfect-information games. More recently, there are some research efforts that combine reinforcement learning with search and have shown its effectiveness in poker games such as heads-up no-limit Texas Hold’em poker and DouDizhu \cite{brown2020combining, jiang2019deltadou}. 

However, due to the complexity of DouDizhu, traditional reinforcement learning methods such as DQN \cite{mnih2015human} and A3C \cite{mnih2016asynchronous} exhibit poor performance in this game as discussed above. Even an improved method, \emph{i.e.} Combinatorial Q-Network, also fails to achieve satisfactory performance. What's more, DeltaDou \cite{jiang2019deltadou}, which infers the hidden information and uses MCTS to combine RL with search, is computationally expensive and depends on human expertise, limiting its practicability and performance. To this end, DouZero \cite{zha2021douzero} utilizes Monte-Carlo methods \cite{sutton2018reinforcement} and manages to defeat all DouDizhu AI programs by now. We note that this technique is also adopted in some other game AIs, such as a modern board game, Kingdomino, and a kind of new chess, Tibetan Jiuqi \cite{gedda2018monte,zhou2021design}. But unlike these environments, DouDizhu is a complex imperfect-information game that requires competition and cooperation over the large state and action space. The amazing performance of DouZero reveals the good results of Monte-Carlo methods in such large-scale complex card games, providing new insight into future research on handling complex action space, sparse reward and imperfect information.

\begin{figure*}[t]
	\centering
	\subfloat[The overall framework]{
		\includegraphics[width=0.63\textwidth]{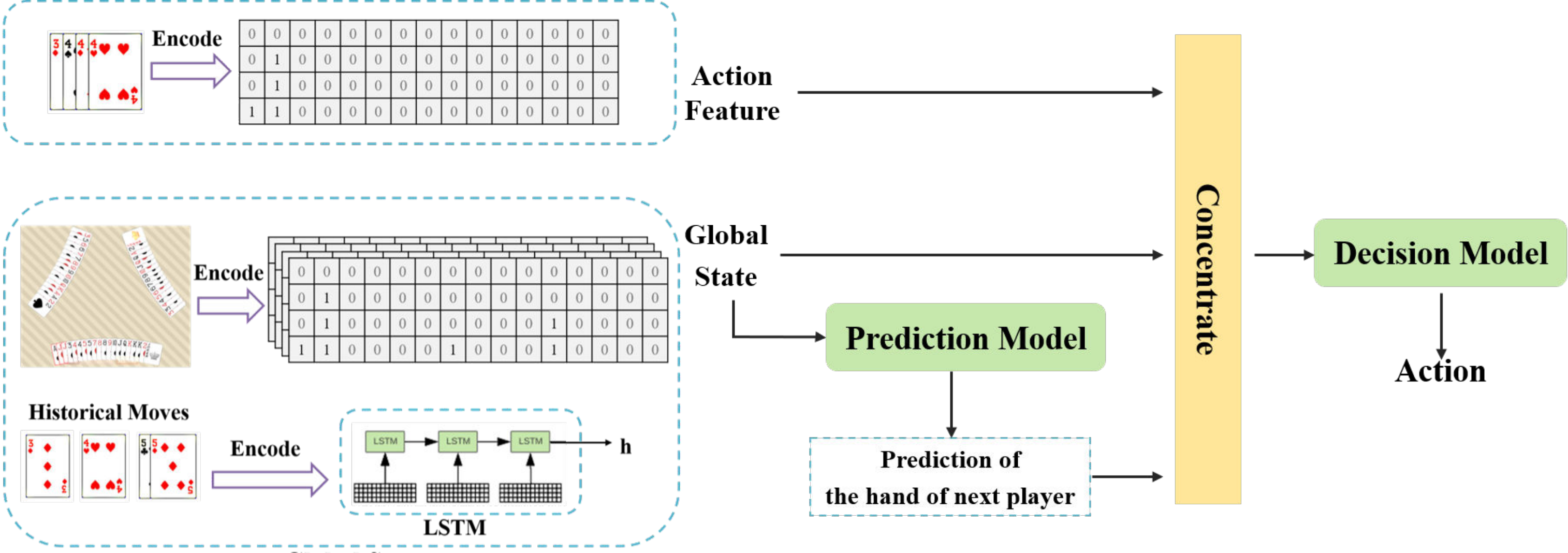}
		\label{fig3-1}
	}
	\subfloat[The details about prediction model]{
		\includegraphics[width=0.33\textwidth]{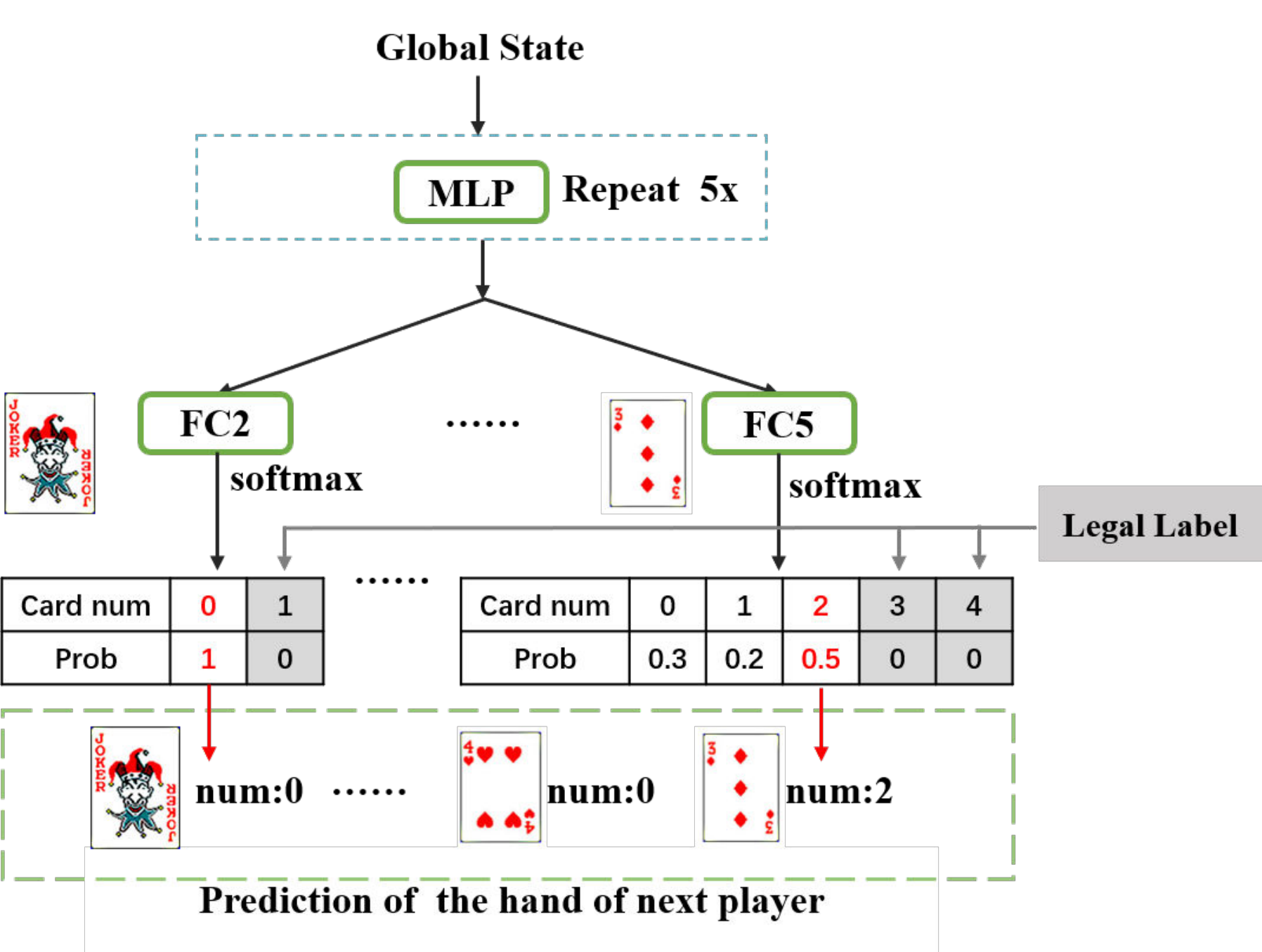}
		\label{fig3-2}
	}
	\caption{An overview of the framework that combines opponent modeling with DouZero and the details about the prediction model. The prediction model takes the state information, which is the same as DouZero, as input and outputs the probability of the number of every card in the hand of the next agent. The decision model is trained using deep Monte-Carlo algorithm like DouZero. The prediction result about hand cards of the next player is concatenated with the state features as well as the action features and all these information is input to decision model to decide which action to take. As for the prediction model, it can be viewed as a multi-head classifier, which consists of a layer of LSTM to encode historical moves, five shared layers of MLP and multi-head FC layers to output the probability. We can extract ``legal label'' from the state information, which represents how many cards of each kind the next player has at most, to help filter out impossible answers.  }
	\label{fig3}
\end{figure*}

\subsection{Opponent Modeling for Games}
In human practice, gaining an abstract description of the opponent will give the player a clear advantage in games, especially imperfect-information games. As a result, opponent modeling has attracted substantial attention in game AI. For example, Southey \emph{et al.} \cite{southey2012bayes} put forward a Bayesian probabilistic model for poker games which infers a posterior over opponent strategies and makes an appropriate response to that distribution. In another complex imperfect-information game, Mahjong, an AI bot is designed based on opponent modeling and Monte Carlo simulation \cite{mizukami2015building}. In this work, the opponent models are trained with expert game records and the bot decides the move using the prediction results and Monte-Carlo simulation. What's more, Schadd \emph{et al.} \cite{schadd2007opponent} propose an approach for opponent modeling in RTS games . It employs hierarchically structured models to classify the strategy of the opponent, where the top-level can distinguish the general style of the opponent and the bottom level can classify the specific strategies that define the opponent's behaviour. 

Recently, inspired by the success of reinforcement learning, many researchers combine opponent modeling with reinforcement learning and have made much progress. In combination with deep Q-learning, opponent modeling achieves superior performance over DQN and its variants in a simulated soccer game and popular trivia game \cite{he2016opponent}. Knegt \emph{et al.} \cite{knegt2018opponent} introduces the opponent modeling technique into an arcade video game using reinforcement learning, which helps the agent predict opponents' actions and significantly improves the agent's performance. In addition, opponent modeling can be adopted in multi-agent reinforcement learning problems where RL agents are designed to consider the learning of other agents in the environment when updating their own policies \cite{foerster2017learning}. Another promising solution is to mimic human players by combining opponent models used by expert players and reinforcement learning \cite{teofilo2012adapting}. All the above works demonstrate that combining opponent modeling with reinforcement learning is beneficial to achieve performance gain in multi-agent imperfect-information games, which also inspires this work.

\section{Preliminary}
In this section, we first discuss the main algorithm of DouZero, \emph{i.e.} Deep Monte Carlo (DMC), which generalizes Monte Carlo (MC) method with deep neural networks for function approximation. Then, we briefly describe the details of DouZero system.
% \subsection{Deep Monte-Carlo}

As a key technique in reinforcement learning, Monte Carlo (MC) method learns value functions and optimal policies from experience, namely, sampling sequences of states, actions and rewards from actual or simulated interactions with the environment \cite{sutton2018reinforcement}. This technique is designed for episodic tasks, where experience can be divided into episodes that eventually terminate, and it updates the value estimation and policy only when an episode is completed. To be specific, after each episode, the observed returns are used for policy evaluation and then the policy can be improved at the visited states in the episode. To optimize a policy $\pi$ using MC methods, the procedure is intuitively described as follows:
\begin{enumerate}
    \item Generate an episode using $\pi$.
    \item For each state-action pair $(s, a)$ visited in the episode, calculate and update $Q(s,a)$ with the average return.
    \item For each state $s$ in the episode, update the policy: $\pi(s) \leftarrow argmax_{\textbf{a} \in \emph{A}}Q(s, \textbf{a} )$.
\end{enumerate}

When putting MC methods into practice, we can utilize epsilon-greedy to balance between exploration and exploitation in Step 1. Also, the above procedure can be naturally combined with deep neural networks, leading to Deep Monte-Carlo (DMC). In this way, the Q-table $Q(s, a)$ can be replaced by neural networks which can be optimized with mean-square-error (MSE) loss in Step 2.

As DouDizhu is a typical episodic task, MC is naturally suitable for this problem. What's more, DMC requires a large amount of experience for training while it's easy to generate data efficiently in parallel, which can also alleviate the issue of variance. In addition, adopting DMC in DouDizhu has some clear advantages compared to other reinforcement learning algorithms, such as policy gradient methods and deep Q-learning, which can be referred to in DouZero \cite{zha2021douzero}. Owing to the advantages that DMC has in DouDizhu, DouZero adopts this algorithm and achieves an outstanding performance.

% \subsection{DouZero}
In the implementation of DouZero system, it makes use of a self-play procedure, where the actors play games to generate samples while the learner updates the network using these data. The input of the network consists of state features and action features. The state feature represents the information that is known to the player, while the action feature describes the legal move corresponding to the current state. Specifically, the action in action features is encoded with a one-hot 4$\times$15 card matrix. For the state features, they contain card matrices that represent the hand cards, the union of other players' hand cards, the played cards of other players and the most recent moves and some one-hot vectors that represent that number of cards of other players, and the number of bombs played so far. 
% When they are input to the network, the encoded state feature is replicated and concatenated with different encoded legal moves. In this way, the network is able to compute the $Q(s,a)$ value for each move under current state, thus the best action can be chosen. 
For the architecture, a layer of LSTM is used to encode historical moves and the output is concatenated with other state/action features. There are six layers of MLP with a hidden size of 512 to produce $Q$ values.

Besides, the system parallelizes DMC with multiple actor processes and one learner process. The learner maintains three global networks for the three positions and updates them to approximate the target values based on data samples generated by actor processes. Each actor maintains three local networks which are synchronized with the global networks periodically. The communication of the learner and actors is implemented with three shared experience buffers. In this way, the system can be trained in an effective self-play procedure.

\section{Method}
% 介绍下蒙特卡洛方法（可借鉴douzero）和对手建模，然后讲一下系统的实现细节
In this section, we introduce opponent modeling and coach network in our design and describe how they are applied.

\begin{figure}[t]
	\centering
	\includegraphics[width=0.98\columnwidth]{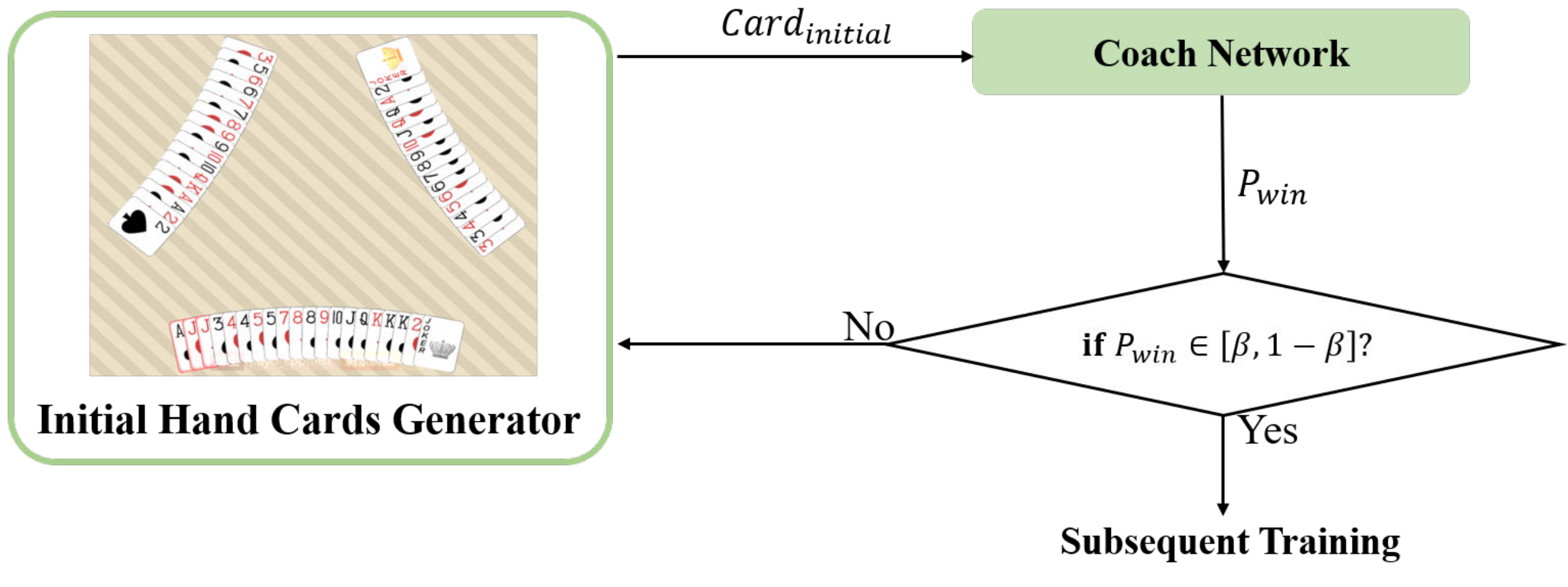}
	\caption{The overview of the framework that utilizes coach network. In this figure, we use the  $Card_{initial}$, $P_{win}$ and $\beta$ to represent generated initial hand cards, the predicted probability of winning for Landlord and the threshold value, respectively. The coach network is composed of one embedding layer and several fully connected layers and the model takes $Card_{initial}$ as input and outputs $P_{win}$. If $P_{win}$ is in the range defined, which is decided by $\beta$, the game with such $Card_{initial}$ will be carried on and generates samples for training. Otherwise, another initial hand cards will be generated.  }
	\label{fig4}
\end{figure}

\subsection{Opponent Modeling}
\label{chap:oppo}
Opponent modeling studies the problem of constructing models to make predictions about various properties of the modeled agents, \emph{e.g.} actions, goals and so on. Classic methods such as policy reconstruction \cite{carmel1998model} and plan recognition \cite{fagan2003case} tend to develop parametric models for agent behaviours. These methods tend to decouple the interactions between the modeled agent and others to simplify the modeling process, which may introduce bias when there exists coupling between agent interactions. In this way, executing opponent modeling when concurrently training all the agents in a self-play procedure is more natural \cite{bansal2017emergent} and suitable to the training procedure of DouDizhu AI system. What's more, concurrent learning helps opponent modeling adapt to different levels of the agent as it has witnessed the evolution of the agent's skills during training.

When adopting opponent modeling in DouDizhu, we predict the hand of the player behind current agent so that the model can make decisions accordingly. As for the implementation of opponent modeling, we can naturally take advantage of deep neural networks to make predictions. To avoid confusion with the network that chooses which move to take, we call the part of opponent modeling as ``prediction model'' and the part that makes decisions as ``decision model''. Following the practice of DouZero that trains three models for the three players in the game, we also train three prediction models for opponent modeling. The prediction model can be viewed as a multi-head classifier and outputs the probability of the number of every kind of card in the hand of the next agent. To be specific, it has to predict how many Card 3, how many Card 4, \emph{etc}, the next player has in his hand. Since the environment of DouDizhu is easy to realize, we can acquire the true hand of the next player and use it as labels to train the prediction model. What's more, taking Card 3 as an example, we can also know how many card 3 of one kind the next player has at most, which can be calculated by the agent's own hand and how many Card 3 has been played. We call this information ``legal label'' and this information can be utilized to help the training of prediction models as it can be used to filter out the wrong answers.

As for the input of prediction models, we make use of the same state features as DouZero. The architecture of prediction models is also similar to DouZero with a layer of LSTM to encode historical moves and five shared layers of MLP. The final layer works as a multi-head classifier where each head corresponds to a fully connected layer and outputs the prediction of one kind of card. This model is trained using cross-entropy loss function. As for the decision model, the features used are also similar to DouZero, except for the prediction of hand cards of the next player in state information. For simplicity, we just concatenate the prediction results as well as original state features for state input of decision models. To sum up, the overview of the framework that combines opponent modeling with DouZero is shown in Figure~\ref{fig3}.

\subsection{Coach-guided Learning}
\label{chap:coach}
During the training of DouDizhu AI system, we discover that the training process costs a lot of time. To this end, we propose a method to help the agent master the skills faster. In this work, our DouDizhu AI system does not have a bidding phase as the bidding network in DouZero is trained with supervised learning. In other words, the initial hand cards of the three players are fixed at the beginning of the game. However, as a shedding-type game where the players' objective is to empty one's hand of all cards before others, the quality of the initial hand cards has a great impact on the result of this game. If one player gets a very strong hand at the beginning, he can win easily as long as he does not make serious mistakes. In this way, such initial cards are of little value for learning as they can hardly help the agent learn new knowledge. On the other hand, if one player always plays matches where the initial hand cards are relatively balanced, he can learn some skills faster and better as he will lose and receive a negative reward if he makes any unsuitable decision. In the setting of DouZero, we uncover that the initial cards of the three players are generated randomly so that quite a few samples may be not matched in strength. However, the actors still have to play the game using these initial cards that are heavily unbalanced, which also takes much time. If we only allow the actors to generate samples that are based on balanced initial hand cards, the agent can learn faster and form policies that can deal with such situation.

Based on the above discussion, we propose a coach network to identify whether the initial hand cards are balanced in strength. It takes the initial hand cards of the three players as input and outputs the predicted probability of winning for the Landlord in one game, which we call $P_{win}$. Then we can set a threshold, which is represented with $\beta$, to filter out the games whose $P_{win}$ is too small or too big. In this case, there is no need for the actors to play with these initial hand cards, thus setting aside time to carry on more valuable matches. 

The input of coach network is the vectors of initial hand cards for Landlord and Peasants, whose dimensions are 20 and 17, respectively. For the architecture of coach network, it consists of an embedding layer to process the input vectors and several layers of fully connected layers to extract representations and make predictions. As our DouDizhu AI system is trained in a self-play manner, the coach network is also concurrently trained with the decision models. The results of self-play games can be used as labels for training the model. Considering that the AI system learns from scratch, the threshold is set to 0 at first and increases through the training process. What's more, we only need to train one coach network for prediction as this module has nothing to do with the positions in DouDizhu. In other words, our coach network only works at the beginning of one game to pick suitable initial data and does not influence the subsequent processes. Therefore such idea can also be transferred into the development of other similar game AIs and benefits the training.

\begin{figure}[t]
	\centering
	\includegraphics[width=1\columnwidth]{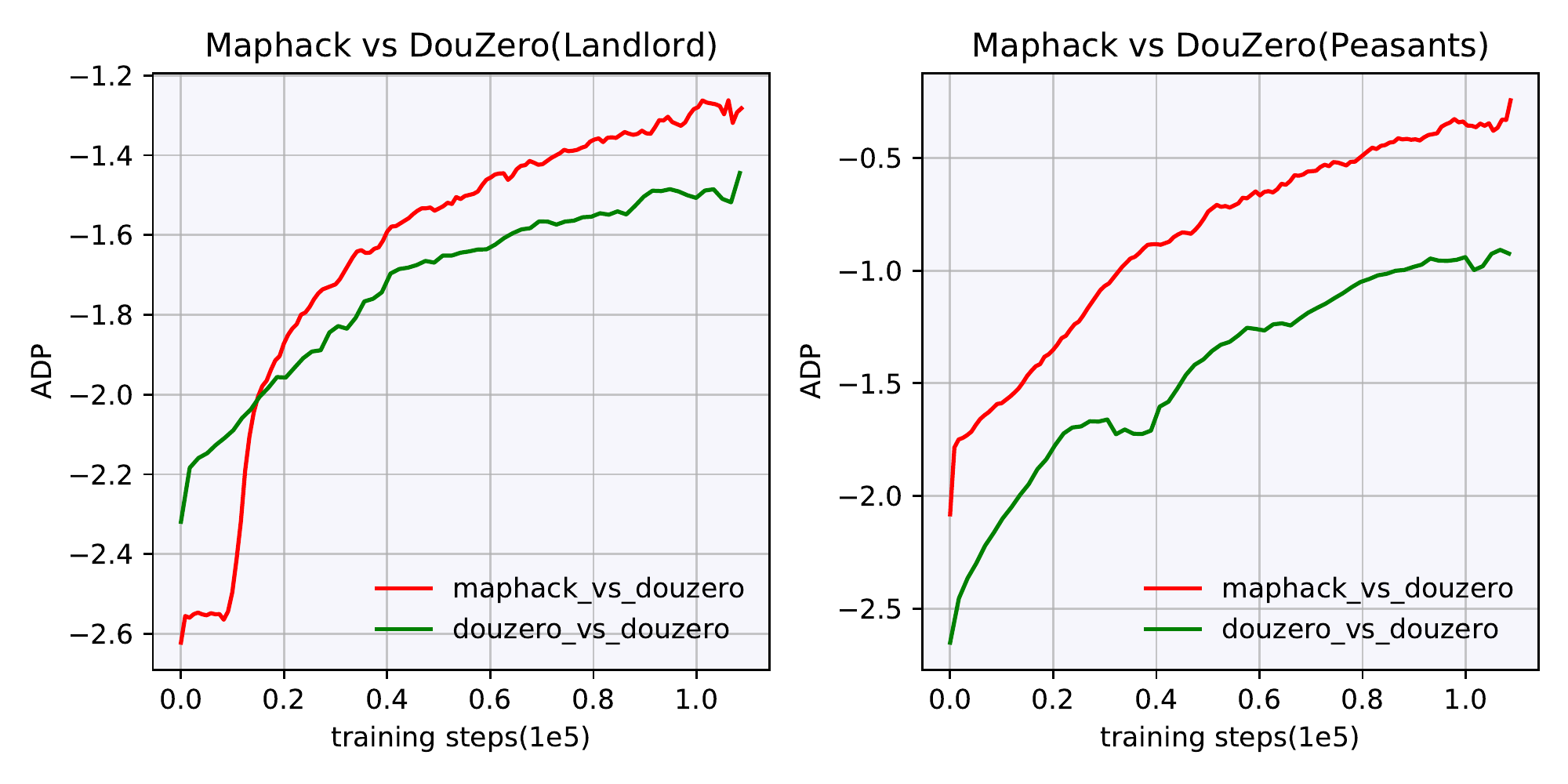}
	\caption{ADP of ``maphack'' models, which can see the hand cards of the next player, and DouZero models. Both these models are tested with DouZero baseline that is trained with ADP. ``Landlord'' means that the models play as Landlord against Peasants of DouZero baseline and the same goes for the reverse.}
	\label{fig5}
\end{figure}

\section{Experiment}
In this section, we conduct experiments to demonstrate the effectiveness of the improvement that we introduce to DouZero. To be specific, we first evaluate the performance of opponent modeling and coach network, respectively, and then combine them together. All experiments are conducted on a server with 4 Intel(R) Xeon(R) Gold 6252 CPU @ 2.10GHz and GeForce RTX 2080Ti GPU. Our codes are available at https://github.com/submit-paper/Doudizhu.

\subsection{Experiment Settings}
Exploitability is a commonly used measure of strategy strength in poker games \cite{johanson2011accelerating}. However, the huge state and action space in DouDizhu make it intractable to calculate exploitability, not to mention that there are three players in this game, which brings more difficulty in evaluation. In order to evaluate the performance of the model, we launch tournaments that include the two opposite sides of Landlord and Peasants, following what DouZero \cite{zha2021douzero}and Deltadou do \cite{ jiang2019deltadou}. To be specific, for two competing algorithms $A$ and $B$, they will first play as Landlord and Peasants, respectively, for a given deck. Then we switch the sides, \emph{i.e.} A takes Landlord position and B takes Peasants position, and they play the same deck again. To show the performance of the model in the training process, we execute the test for 10000 episodes every 30 minutes. As our DouDizhu AI is based on DouZero, we just compare the performance between them. We make use of the open-source models of DouZero as the opponent. To demonstrate the improvement, we also realize the original DouZero to intuitively exhibit the performance difference. As for the evaluation metrics, we also follow DouZero and use Winning Percentage (WP) and Average Difference in Points (ADP). Specifically, WP represents the number of games won by algorithm $A$ divided by the total number of games. ADP indicates the average difference of points scored per game between algorithm $A$ and $B$, where the base point is 1 and each bomb will double the score. 

Our implementation is based on DouZero and training schedules such as the number of actors and training hyperparameters are kept the same as the default ones. As the DouDizhu environment is realized by ourselves, the reward also needs to be defined. The evaluation metrics of WP and ADP can be utilized when defining the reward. For WP, the agent winning a game is given +1 reward otherwise -1 reward while ADP can be directly used as rewards for ADP settings. DouZero provides two kinds of models which are trained using WP and ADP, respectively. For simplicity, we train our AI system with ADP as objective and compare its performance with the corresponding baseline. Also, we use the metric of ADP when evaluating the performance of the models.

\begin{figure}[t]
	\centering
	\includegraphics[width=1\columnwidth]{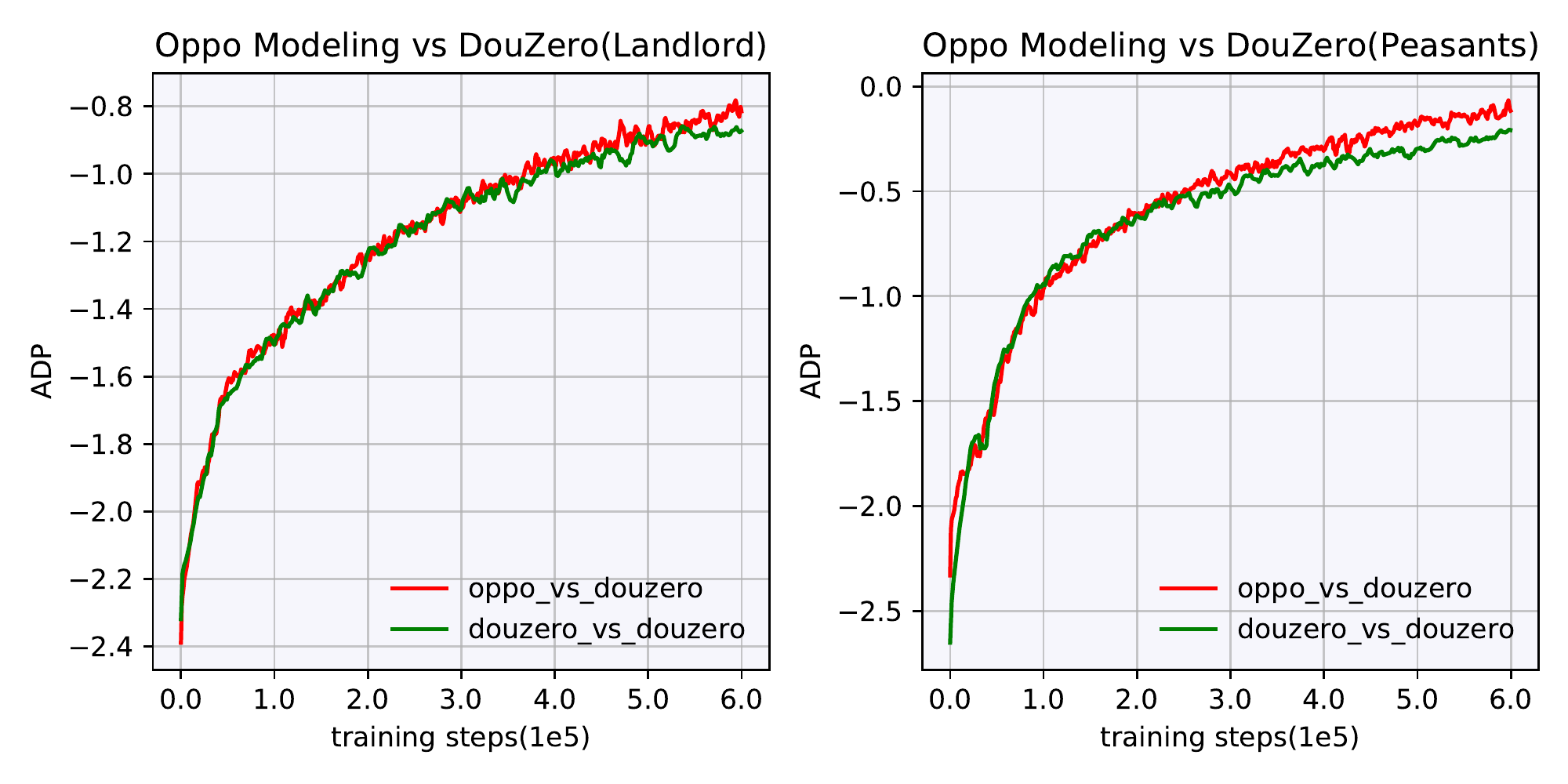}
	\caption{ADP of models, which combine opponent modeling and DouZero, and DouZero models. Both these models are tested with DouZero baseline that is trained with ADP. ``Landlord'' means that the models play as Landlord against Peasants of DouZero baseline and the same goes for the reverse. }
	\label{fig6}
\end{figure}

\subsection{Evaluation on Opponent Modeling}
In this part, we demonstrate the effectiveness of introducing opponent modeling to DouDizhu. As the state features utilized by DouZero contain all the information that can be known, the information about the hand cards of the next player is included implicitly while the idea of opponent modeling is essentially making such information explicit. In order to investigate whether such an idea helps the agents learn better, we firstly make a pre-experiment where we add the hand cards of the next player into state features directly, whose result is shown in Figure~\ref{fig5}. It can be observed that adding the hand cards of the next player into state features indeed boosts the performances of the agents, especially for Peasants. We assume that the obvious improvement of Peasants is attributed to the fact that knowing the hand cards of the next player helps Peasants not only choose cards that the Landlord can't afford but also cooperate with the teammate better. Whereas for the Landlord, knowing the hand cards of next player indeed helps to make decisions, but if the hand is weak, even having such information can not help a lot. To sum up, the result of the pre-experiment illustrate that introducing explicit representations of the next player's hand cards improves the performance of DouDizhu AI.

After verifying the validity of our idea, we concurrently train the prediction models as well as the decision models as is discussed in Section~\ref{chap:oppo} and the result is shown in Figure~\ref{fig6}. It reveals that introducing opponent modeling to DouZero mainly improves the performance of models of Peasants, which is corresponding to the analysis above. Although the models perform worse than DouZero at first because the network has to take more features as input and has more neurons, which will slow down learning, they manage to grasp more knowledge after enough training and achieve a performance better than DouZero.

\begin{figure}[t]
	\centering
	\includegraphics[width=1\columnwidth]{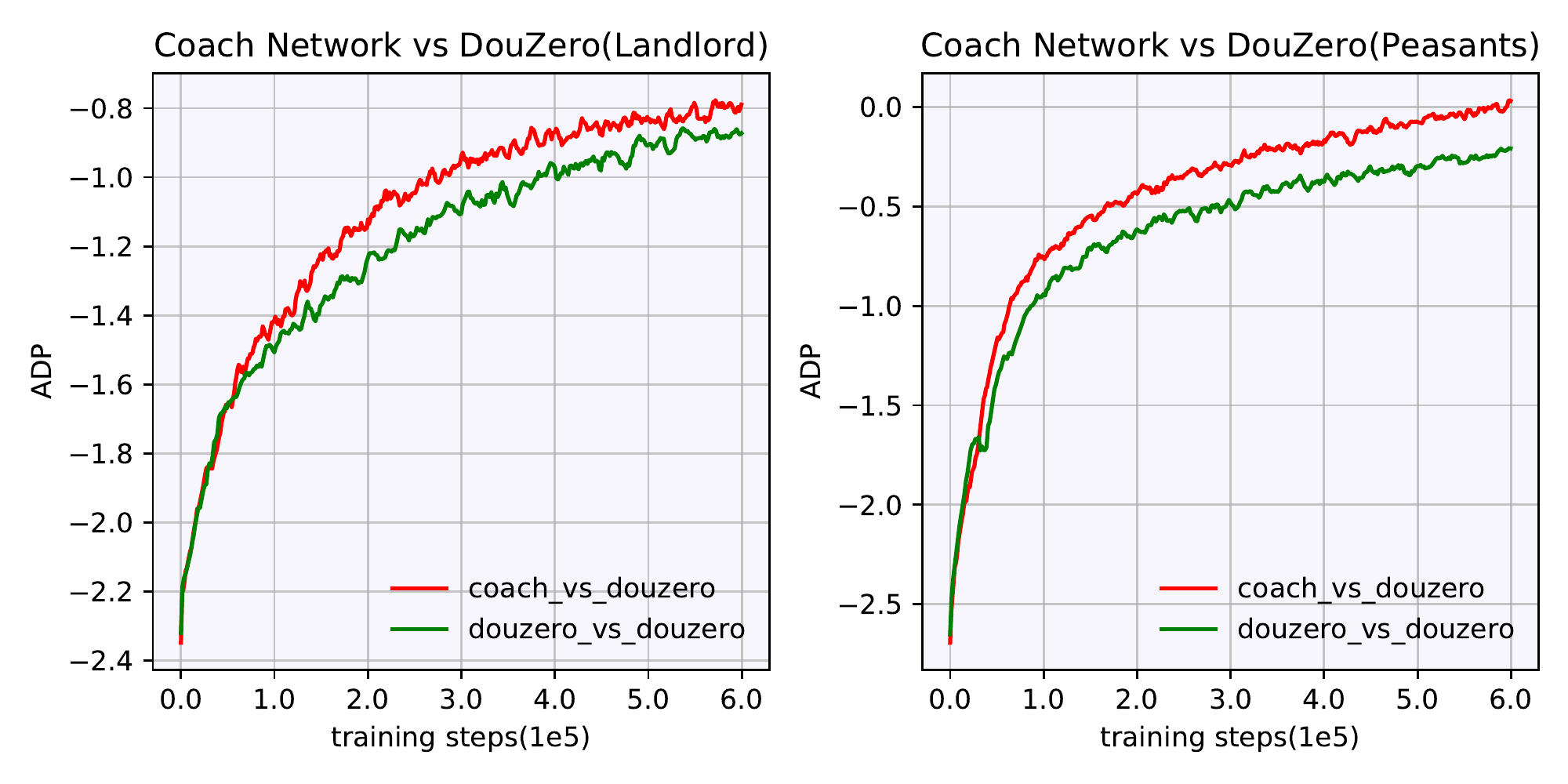}
	\caption{ADP of models, which combine coach network with DouZero, and DouZero models. Both these models are tested with DouZero baseline that is trained with ADP. ``Landlord'' means that the models play as Landlord against Peasants of DouZero baseline and the same goes for the reverse. }
	\label{fig7}
\end{figure}

\begin{table*}[htbp]
    \centering
    \begin{threeparttable}
        \begin{tabular}{ c|c c c c c }
    		\toprule[1pt]
            & Landlord & Landlord\_down & Landlord\_up &  Prediction of $P_{win}$ for Landlord & Actual result(Landlord) \\
            \midrule
            \rule{0pt}{8pt} %改变行高
            Case1 & 3455677789JQKAAAA22R  & 334569TTTJJQQQKK2 & 344566788899TJK2B & 0.9932 & Win\\ 
            \rule{0pt}{8pt} %改变行高
            Case2 & 45667788889TTTKKA22B  & 334567TJJJQQQQK22 & 33445567999JKAAAR & 0.1726 & Lose\\
            \rule{0pt}{8pt} %改变行高
            Case3 & 3455556677799JJQKAAB  & 3467889TTQKKK222R & 33446889TTJJQQAA2 & 0.5843 & Lose\\
            \bottomrule
        \end{tabular}
        % \begin{tablenotes}    %这行要添加， 从这开始
        %     \footnotesize               %这行要添加
        %     \item[1] 
        %     % The RL agent without satisfying zero-sum co
        %     \item[2] The cards are ranked by Red Joker, Black Joker, 2, A, K, Q, J, 10, 9, 8, 7, 6, 5, 4, 3
        % \end{tablenotes}  
    \end{threeparttable}
    \caption{Case study to show the effect of ``coach network''. It predicts the winning probability of Landlord based on the initial hand cards of the three players. We pick some cases from games from Botzone to show the predicted results of ``coach network'' and also show the actual result from the view of the Landlord. To be mentioned, T means 10, J means Jack, Q means Queen, K means King, A means Ace, B means Black Joker, and R means Red Joker.}
    \label{tab1}
\end{table*}

\subsection{Evaluation on Coach Network}
Apart from the experiments above, we also conduct experiments to show how coach network'' performs in DouDizhu game. The training procedure is discussed in Section~\ref{chap:coach} and the upper limit of threshold $\beta$ is set to be 0.3. The result of the experiment is shown in Figure~\ref{fig7} and the significant improvement proves the effectiveness of this method. It can be observed that the improvement of Peasants is also greater than that of Landlord. Considering that Peasants have an advantage in this game due to cooperation, this phenomenon is acceptable as they can learn more skills in balanced games. Besides, this coach-guided learning strategy just controls the initial state of the game while the results demonstrate the significant improvement it can bring. This fact reveals that the luck factor plays an important role in such kind of imperfect-information games. In other words, our method can be migrated into other environments, helping game AI achieve better performance.

What's more, we also show some cases about the prediction of our coach network from games on Botzone platform, which is illustrated in Table~\ref{tab1}. In case 1, it can be observed that the Landlord is allocated with a very strong hand, which consists of most cards of high rank and cards of low rank that can compose other combinations so that the Landlord can win the game easily. As for case 2, even Landlord has a bomb in his hand, the hand cards of Peasants are also very strong. What's worse, the Landlord also has quite a few cards of low rank that are difficult to play out. In case 3, the initial hand cards are relatively balanced. However, the Peasant win the game finally, indicating the importance of cooperation. This example illustrates that the balanced samples can indeed help the agents learn cautious policy and cooperation better, thus proving the correctness of our idea.

\subsection{Combination of Two Methods}
\begin{figure}[t]
	\centering
	\includegraphics[width=1\columnwidth]{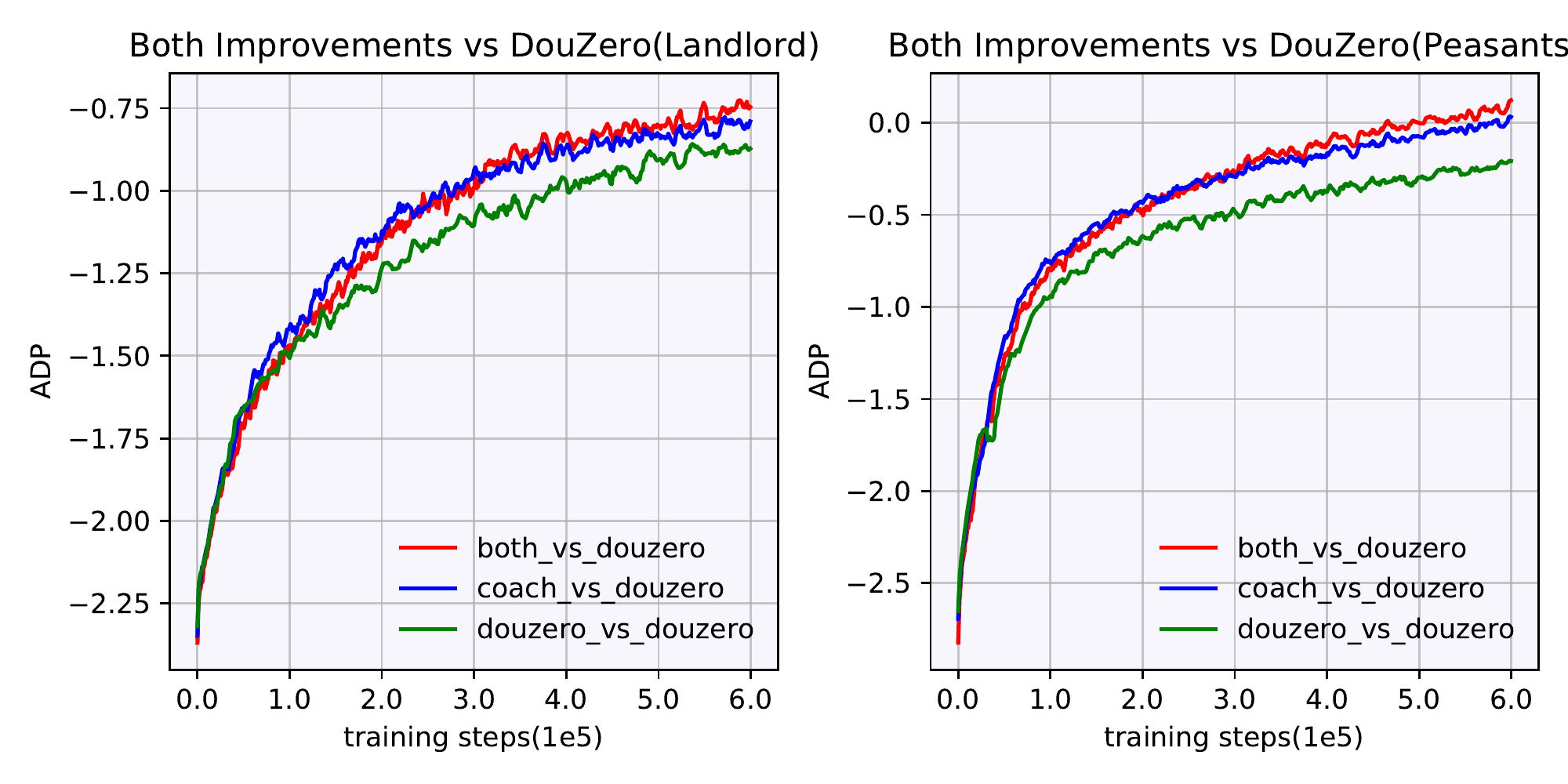}
	\caption{ADP of models, which combine both improvements with DouZero, and DouZero models. Both these models are tested with DouZero baseline that is trained with ADP. ``Landlord'' means that the models play as Landlord against Peasants of DouZero baseline and the same goes for the reverse. For comparison, the result of models improved by ``coach network'' is also included.}
	\label{fig8}
\end{figure}

From the above discussion, it is known that both our improvements can help enhance the performance of DouZero. The result of combining these two methods is shown in Figure~\ref{fig8}. As the improvement of ``coach network'' is more obvious than opponent modeling, to intuitively demonstrate whether the combination of these two techniques brings further improvement, we also add the result of just using ``coach network'' in the figure. It can be observed the performance is a little worse than just using coach network at first, which is consistent with the discussion of just introducing opponent modeling. To be mentioned, when the performance of the models reaches a certain level, achieving a little improvement is very difficult so the progress that combining the two methods makes is not that apparent. However, further improvement still proves the effectiveness of combination of the two methods.

To comprehensively compare the performance of our DouDizhu AI, we upload our final model to BotZone \cite{zhou2018botzone}, an online platform with DouDizhu competition. This platform supports more than 20 games apart from DouDizhu, including Go, Mahjong, Chess and so on. There are more than 3500 users on this platform uploading their bot programs to compete with other bots in a selected game. Botzone maintains a leaderboard for each game, which ranks all the bots in the Botzone Elo system by their Elo rating scores. In the Botzone Elo of DouDizhu (named ``FightTheLandlord'' on the platform), each game is played by two bots, with one acting as the Landlord and the other as Peasants. A pair of games are played simultaneously where the two bots play different roles and the initial hand cards also keep unchanged. Although Elo rating is generally considered as a stable measurement of relative strength, DouDizhu Elo ranking on Botzone suffers from some fluidity due to the nature of high variance of this game. What's more, due to the limit of server resources, Elo rating games are not scheduled very frequently. One bot plays less than 10 Elo rating games on average every day so that it may take a lot of time to achieve a stable ranking. However, keeping a high ranking can still prove the strength of one AI system. Even if DouZero has obvious superiority over other DouDizhu AI systems trained by reinforcement learning, it has ranked about 20th so far on Botzone leaderboard as most bots are realized by strong heuristic rules. Nonetheless, our DouDizhu AI has always ranked top five, even ranked first for several months, proving the effectiveness of the improvements that we make.

\section{Conclusion and Future Work}
In this work, we put forward some improvements to the state-of-the-art DouDizhu AI program, DouZero. Inspired by the human player's prediction about opponents' hand cards in practice, we introduce opponent modeling. Based on the nature of high variance of this game, we originally propose a ``coach network'' to pick valuable samples to accelerate the training. The outstanding performance of our AI on the Botzone platform proves the effectiveness of our improvement.

Although our DouDizhu AI performs well after adopting these techniques, there is still room for improvement. First, to better show the effect of our improvement, we do not make changes on the architectures of neural networks in DouZero unless necessary. We plan to try other neural networks such as convolutional neural networks like ResNet \cite{he2016deep}. Second, we find that there are still some cases where the model can not make good decisions. We hope to combine search with our AI to enhance the performance as search plays an important role and performs well in research about game AI \cite{bouzy2020recursive, ariyurek2020enhancing}. Finally, we will investigate how to improve the sample efficiency with experiment replay \cite{zhang2017deeper} as it still costs a lot of time even utilizing our ``coach network''. In addition, we will also try to transfer our methods to other games for stronger game AIs.
\bibliographystyle{IEEEtran}
\bibliography{refer.bib}

\end{document}